\documentclass[10pt,twocolumn,letterpaper]{article}

\usepackage{wacv}
\usepackage{times}
\usepackage{epsfig}
\usepackage{graphicx}
\usepackage{amsmath}
\usepackage{amssymb}

\usepackage{multirow}
\usepackage{lipsum}
\usepackage{subcaption}
\usepackage{makecell}
\usepackage{booktabs}
\usepackage{times}

\usepackage{textgreek}
\usepackage{siunitx}
\DeclareSIUnit{\pp}{\textup{p.p.}}
\usepackage{tabularx}
\usepackage{algorithmic}
\usepackage{algorithm}
\usepackage[pagebackref=true,breaklinks=true,letterpaper=true,colorlinks,bookmarks=false]{hyperref}
\usepackage[accsupp]{axessibility}  

\def\wacvPaperID{1279} 

\wacvfinalcopy 

\ifwacvfinal
\fi

\ifwacvfinal
\usepackage[breaklinks=true,bookmarks=false]{hyperref}
\else
\usepackage[pagebackref=true,breaklinks=true,colorlinks,bookmarks=false]{hyperref}
\fi

\begin{document}

\title{Data Augmented 3D Semantic Scene Completion with 2D Segmentation Priors}

\author{Aloisio Dourado, Frederico Guth ~ and ~ Teofilo de Campos \\
University of Brasilia\\
Campus Darcy Ribeiro. Asa Norte, Brasilia, DF - 70910-900, Brazil\\
{\tt\small aloisio.dourado.bh@gmail.com} ~
{\tt\small fredguth@fredguth.com} ~
{\tt\small t.decampos@oxfordalumni.org}
}

\twocolumn[{%
	\maketitle
	\renewcommand\twocolumn[1][]{#1}%
}]

\ifwacvfinal
\thispagestyle{empty}
\fi

\begin{abstract}

Semantic scene completion (SSC) is a challenging Computer Vision task with many practical applications, from robotics to assistive computing. Its goal is to infer the 3D geometry in a field of view of a scene and the semantic labels of voxels, including occluded regions.
In this work, we present SPAwN, a novel lightweight multimodal 3D deep CNN that seamlessly fuses structural data from the depth component of RGB-D images with semantic priors from a bimodal 2D segmentation network.
A crucial difficulty in this field is the lack of fully labeled real-world 3D datasets which are large enough to train the current data-hungry deep 3D CNNs. In 2D computer vision tasks, many data augmentation strategies have been proposed to improve the generalization ability of CNNs. 
However those approaches cannot be directly applied to the RGB-D input and output volume of SSC solutions. In this paper, we introduce the use of a 3D data augmentation strategy that can be applied to multimodal SSC networks. 
We validate our contributions with a comprehensive and reproducible ablation study. Our solution consistently surpasses previous works with a similar level of complexity.


\end{abstract}

\section{Introduction}
\begin{figure*}
\centering  

\includegraphics[width=\linewidth,trim={25 255 20 2},clip] {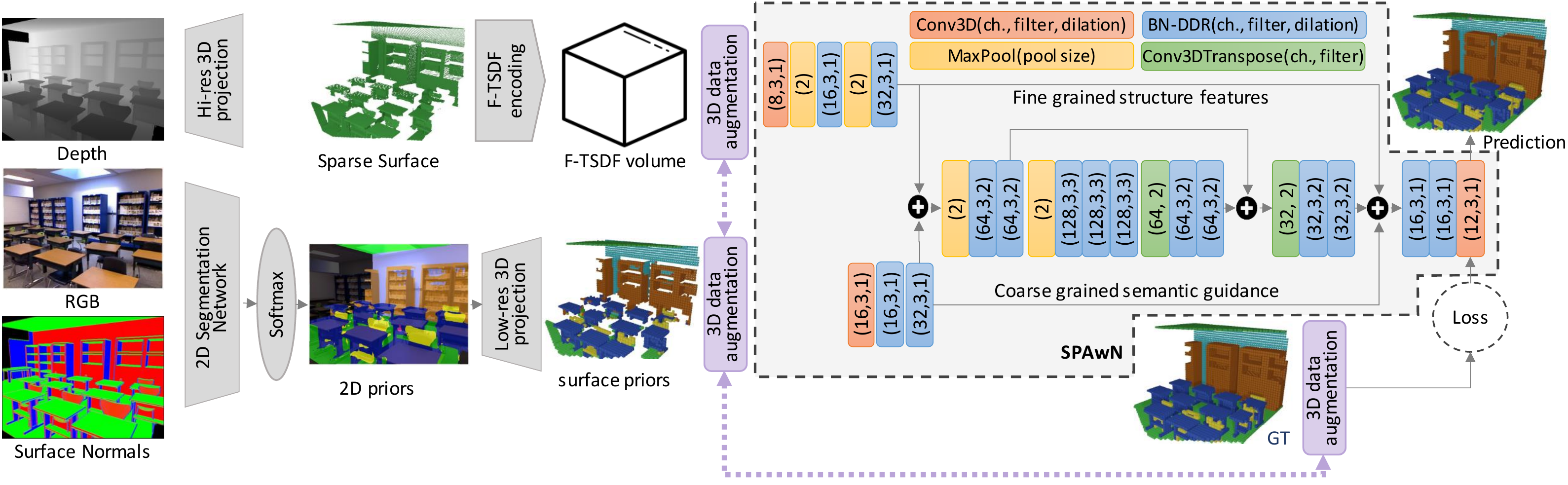}
\caption{\textbf{Overview of our solution}. Our system comprises SPAwN, a 3D CNN that uses 2D priors as semantic guidance, and a novel 3D data augmentation approach for regularization and overfitting reduction. The 2D segmentation network is multimodal, combining RGB and surface normals.
%
(Best viewed in color.)
}
\label{fig:arch}
\end{figure*}

Reasoning about scenes in 3D is a natural human ability that remains a challenge for Computer Vision. 
In the past, the two most common scene understanding tasks were scene completion~\cite{firman_2016} and semantic labeling of visible surfaces~\cite{Gupta_2013, Qi_2017, ren_2012}. 
%
Noticing that these are intertwined tasks, in 2017, Song et al.\ \cite{song_semantic_2017} introduced the Semantic Scene Completion (SSC) task for simultaneously completing occluded voxels and inferring their semantic labels and proposed SSCNet, achieving better results than dealing with these tasks separately. Early approaches only used depth information, ignoring the RGB channels \cite{song_semantic_2017,guo_view-volume_2018}. 
The use of color channels was introduced later \cite{guedes_semantic_2018}.

We present a new approach for exploring information from the RGB-D input (explained in section \ref{sec:proposed_solution}), as~shown in Figure \ref{fig:arch}.
The solution uses 2D prior probabilities from a bimodal 2D segmentation network as semantic guidance to the depth map's structural data. The proposed multimodal 3D network, \emph{SPAwN}, uses a new memory-saving batch-normalized dimensional  decomposition  residual  building  block (BN-DDR) and can be trained on a single 10Gb GPU with a 4 scene mini-batch. 

To overcome the limitations imposed by the lack of sizeable real-world datasets, we are the first to apply 3D data augmentation for the SSC task. Data augmentation is widely used in the training of 2D deep CNNs \cite{NIPS2012_Krizhevsky,ResNet} and its goal is to reduce overfitting by artificially increasing the variety of samples in the training dataset using transformations like flipping, cropping, rotation and color transforms. However, those transformations can not na\"ively be used in 3D applications like semantic Scene completion because of the difference in the number of dimensions of the input (2D) and output (3D). 
In this paper, we propose to apply data augmentation to inner 3D volumes of the solution with three fast 3D transformations in voxel space that preserve the main characteristics of the scene. Our proposed data augmentation approach reduces overfitting and achieves unprecedented levels of semantic completion when compared to previous works of similar memory footprint and complexity. 

We evaluated our contributions with and without pretraining on synthetic data and observed that our method surpasses, by far, all previous state-of-the-art results on both scenarios.  We demonstrate the benefits of the proposed architecture and the data augmentation approach separately, with several experiments in a comprehensive and reproducible ablation study. 
Regarding the proposed augmentation scheme, we evaluate it for training (regular data augmentation) and test (test-time data augmentation). 

 Supplementary material provides additional graphs and data regarding all experiments. All models and training code necessary to reproduce our results and the ablation experiments are publicly available\footnote{Source code:\scriptsize{ 
 \url{https://cic.unb.br/~teodecampos/aloisio/}
 }} .

Our contributions are listed below.
\begin{itemize}
	\item \emph{SPAwN}, a novel lightweight multimodal 3D SSC CNN architecture that uses 2D prior probabilities from a 2D segmentation network. These priors are used as semantic guidance to the structural data from the depth part of the RGB-D input. This architecture can be efficiently trained on a single 10Gb GPU and achieves state-of-the-art results on both real and synthetic data.
    \item \emph{BN-DDR}, a memory-saving batch-normalized dimensional decomposition residual building block for 3D CNNs. It preserves previous approaches' regularization characteristics while consuming much less memory during training.
	\item 
	We are the first to apply a data augmentation technique for 3D semantic scene completion. Our method uses three 3D data transformations which operate on batches directly in GPU tensors. 
	\item \emph{SPAwN} surpasses by far all known previous works with similar memory footprint and pipeline complexity on both real and synthetic data. When trained with the proposed data augmentation strategy, our solution achieves unprecedented levels of SSC scores.
\end{itemize}

\section{Related Work}

\textbf{2D RGB-D Semantic Segmentation} extends regular 2D RGB semantic segmentation with the addition of 3D depth maps obtained from devices such as structured light sensors (\eg Kinect), time-of-flight IR sensors, or stereo cameras. Early works \cite{ren_2012,Gupta_2013,silberman_2012} relied on handcrafted features further fed into a classification model. Later, following the boom of CNNs, models started using the depth map as a fourth input channel \cite{Eigen_2015_ICCV}, eventually encoded in HHA (horizontal disparity, height above ground, and angle with gravity) \cite{Gupta_2014}, before feeding into the neural network \cite{FCNN, Qi_2017, Wang_2018_ECCV}. Lin et al.\ \cite{Lin_2017_CVPR} introduced the RefineNet module to fuse skip-connections from different scales of the segmentation encoder, and \cite{Park_2017_ICCV} extended it to deal with multiple modes introducing the Multimodal Fusion (MMF) module. Later, \cite{Jiao_2019_CVPR} proposed an encoder-decoder network that uses depth maps as ground truth to extract depth embeddings that are later fused to semantic features from the RGB image, thus eliminating the need for depth maps during inference and also achieving the current state-of-the-art in RGB-D semantic segmentation. Our 2D network (explained in detail in section \ref{sec:2d_segmentation_network}) is a simplified version of \cite{Park_2017_ICCV} which did not have the ambition of being state-of-the-art for this auxiliary task. Its design prioritized keeping a low memory footprint.

\textbf{Semantic Scene Completion} was introduced relatively recently \cite{song_semantic_2017} and is already an active field of study in computer vision. Early solutions  \cite{song_semantic_2017, zhang_semantic_2018,guo_view-volume_2018} used only the depth maps of the RGB-D images, neglecting the RGB channels. Guedes \etal\cite{guedes_semantic_2018} reported preliminary results obtained by adding color to an SSCNet~\cite{song_semantic_2017}-like architecture. After that, many solutions \cite{Garbade_2019_CVPR_Workshops,See_and_think_2018,Li_2020_CVPR,attention_2020} started using RGB channels as input to a 2D segmentation network, whose generated features were fed into a 3D CNN. Current state-of-the-art solutions include using RGB edges to improve detection of hard-to-detect classes \cite{edgnet}, using 360$^o$ panoramic images \cite{Dourado_etal_EdgeNet360_VISAPP2020}, fusing multi-scale context information \cite{CCPNet},  multiple separate generators \cite{ForkNet}, and semi-supervision from depth maps \cite{Sketch}. 
All previous methods use a straightforward training pipeline, where the input flows through the network in a single direction, without any loops. More recently, \cite{Cai_2021_CVPR} introduced the scene-instance-scene pipeline, which includes a sequence of semantic segmentation and instance completion networks. This pipeline is iteratively executed multiple times, achieving great results at the cost of requiring much more computational power.

In this paper we present a lightweight and straightforward solution that, like some other methods, explores 2D segmentation. However, we propose a completely novel way of extracting knowledge from the RGB-D channels. Our results are a significant improvement \wrt the state-of-the-art on benchmark datasets for this kind of pipeline and are comparable with the much more expensive iterative solution.

\textbf{Data Augmentation} is a regularization technique that is vastly used in 2D segmentation problems \cite{NIPS2012_Krizhevsky,ResNet,Lin_2017_CVPR, Jiao_2019_CVPR} and other 2D computer vision applications. Regularization and consequent reduction in overfitting is achieved by artificially enlarging the training dataset randomly applying transformations like flipping and cropping to the input images \cite{NIPS2012_Krizhevsky}.  More recently introduced data augmentation strategies include blocking-based approaches like Blockout \cite{Blockuot2016} and Random Erasing \cite{Zhong_Zheng_Kang_Li_Yang_2020}.  
Applying the transformations to the input images is enough for image classification tasks since they do not affect the corresponding labels.
However,  for semantic segmentation tasks, if the transformations affects objects' position in the input image, it is necessary to apply the same transformations to the ground truth maps to preserve the pixel-to-pixel correspondence \cite{NIPS2012_Krizhevsky}. The need for applying the same transformations to the input and the ground truth 
is an extra difficulty
for using data augmentation in RGB-D to voxel segmentation tasks like semantic scene completion. Keeping the correspondence between the input and ground truth representation is necessary, but there is no direct pixel to voxel mapping from the input to the output. When the solution uses some kind of Truncated Sign Distance Function like TSDF or F-TSDF \cite{song_semantic_2017} the problem is exacerbated because changing one pixel of the input would propagate in 3D space through the TSDF-encoded volume in a pyramid shape, affecting a large region. 

In this paper, we overcome those side-effects 
by applying the data augmentation transformations
to the inner 3D volumes of the input and the ground truth volume, rather than the RGB-D channels.
They, therefore, preserve the TSDF or F-TSDF encoding and lead to impressive results. 
     

\section{Proposed Solution}
\label{sec:proposed_solution}


The overall multimodal solution is shown in Figure \ref{fig:arch}. 
Initially, we feed two modes of the input RGB-D image (RGB and surface normals) into a 2D segmentation CNN. Then, we submit the output of the 2D network to a Softmax function and obtain the prior probabilities that will be further projected to a low-resolution 3D voxel volume.
The depth map, a third input mode, is projected to a high-resolution 3D volume and encoded with F-TSDF~\cite{song_semantic_2017}. 
Data augmentation is applied directly to the 3D volumes, including ground truth, before feeding our SPAwN CNN.  
The input branches of SPAwN match the scale and the volumes are fused with an early/late fusion network to produce the final predictions. 

\subsection{2D Segmentation Network}
\label{sec:2d_segmentation_network}
\begin{figure}[t]
\centering  
\includegraphics[trim=2 213 452 0, clip,width=\linewidth] {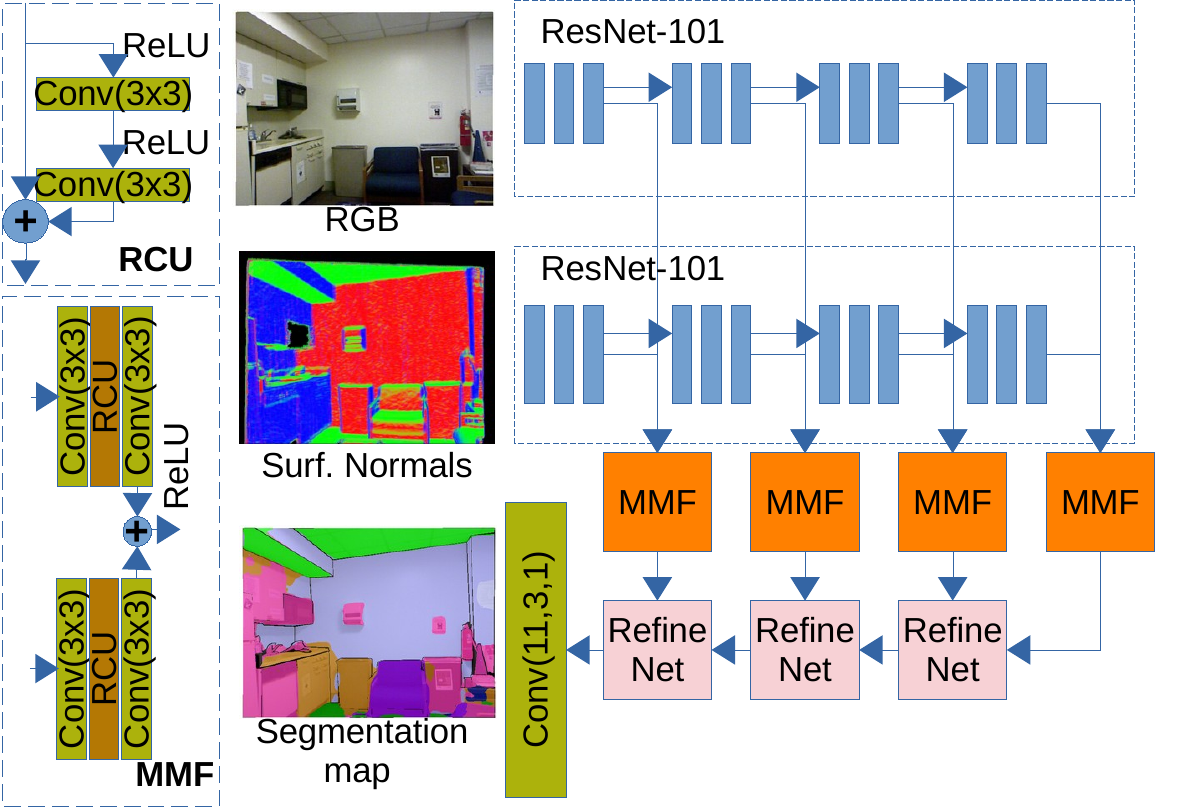}
\caption{\label{fig:2d_network}\textbf{2D bimodal segmentation network architecture.} The Residual Convolution Unit (RCU) and the RefineNet module were first defined in \cite{Lin_2017_CVPR}. Here, we use a simplified MMF block~\cite{Park_2017_ICCV}. (Best viewed in color.)}
\end{figure}

To acquire high-quality 2D priors while keeping the memory footprint low, we designed a tailor-made version of RDFNet \cite{Park_2017_ICCV}. We use a bimodal encoder-decoder 2D RGB-D segmentation network with two ImageNet pre-trained ResNet-101 \cite{ResNet} backbones, one for each input mode, as presented in Figure \ref{fig:2d_network}. The main adjustments to the original RDFNet are simplifications to the MMF module reducing the number of convolution layers; the usage of 3 RefineNet modules, instead of 4;  and the modification of the number of channels of the last layer since we need a classifier for 11 classes (in 2D, the empty or void class is ignored and the original RDFNet was trained for 40 classes).

This customized RDFNet takes two input images:  the color channels of the RGB-D input, as in the original version, and surface normals, instead of the HHA encoded image. The surface normals are obtained from the depth map after aligning the scene to the Cartesian axes following the Manhattan assumption, as in \cite{silberman_2012}. 
Each axis is mapped to one RGB channel, and the absolute normal values are normalized from 0 to 255.  Our goal with this simplified 2D network is only to test the hypothesis that 2D segmentation priors would improve the overall result. 




\subsection{SPAwN Semantic Scene Completion Network}
Our Segmentation Priors Aware Network (\emph{SPAwN}) is a novel 3D CNN  architecture that uses 2D prior probabilities from a semantic segmentation network to guide the depth map's structural information.

\textbf{Depth map projection and encoding.} Following previous works on SSC, we align the scene to the Cartesian axes considering the Manhattan assumption, project the depth map to 3D and then apply F-TSDF \cite{song_semantic_2017} to reduce data sparsity. The dimensions of the 3D space are set to 4.8m horizontally, 2.88m vertically, and 4.8m in depth. Voxel grid size is 0.02m, and the dimension of the resulting volume is $240\times144\times240$. The F-TSDF truncation value is 0.24m. Examples of a projected depth map are shown in Figure \ref{fig:surf}.

\textbf{2D prior probabilities projection and ensemble.} Projecting the output of the 2D network to 3D is a delicate task. One could be tempted to project the features to 3D at the same resolution of the structural volume. However, it would lead to 11 volumes of $240\times144\times240$ voxels, consuming too much GPU memory. 
Our solution, instead, consists in projecting the data from 2D at a lower resolution. We use $60\times36\times60$, the same as the output of the network. We compensate the reduction in details, which come from the structural branch, with a boost in accuracy during downsampling by using a simple yet effective classifier ensemble method known as the ``sum rule'' \cite{Kittler_1998}, as follows. Firstly we obtain the probabilities for the 11-class output by applying a Softmax function, resulting in 11 planes with the exact resolution as the input image. Then, we project each pixel to 3D at low resolution (voxel size = 0,08m). When more than one pixel falls into the same voxel, we sum the probabilities of each class and, to normalize the resulting priors, we divide the probabilities by the number of pixels that fell into the voxel.
%

The previous approach only provides information for the surface voxels. To provide information to non-surface voxels, we add an extra channel for the empty class, and for all non-surface voxels, we set the probability $1$ for voxels belonging to the class ``empty'' and $0$ to the other $11$ classes. An example of projected prior volume compared to the ground truth is shown in Figures \ref{fig:priors} and \ref{fig:gt}.

\begin{figure}[t]
\centering
\includegraphics[trim=0 240 450 0, clip, width=.7\linewidth]{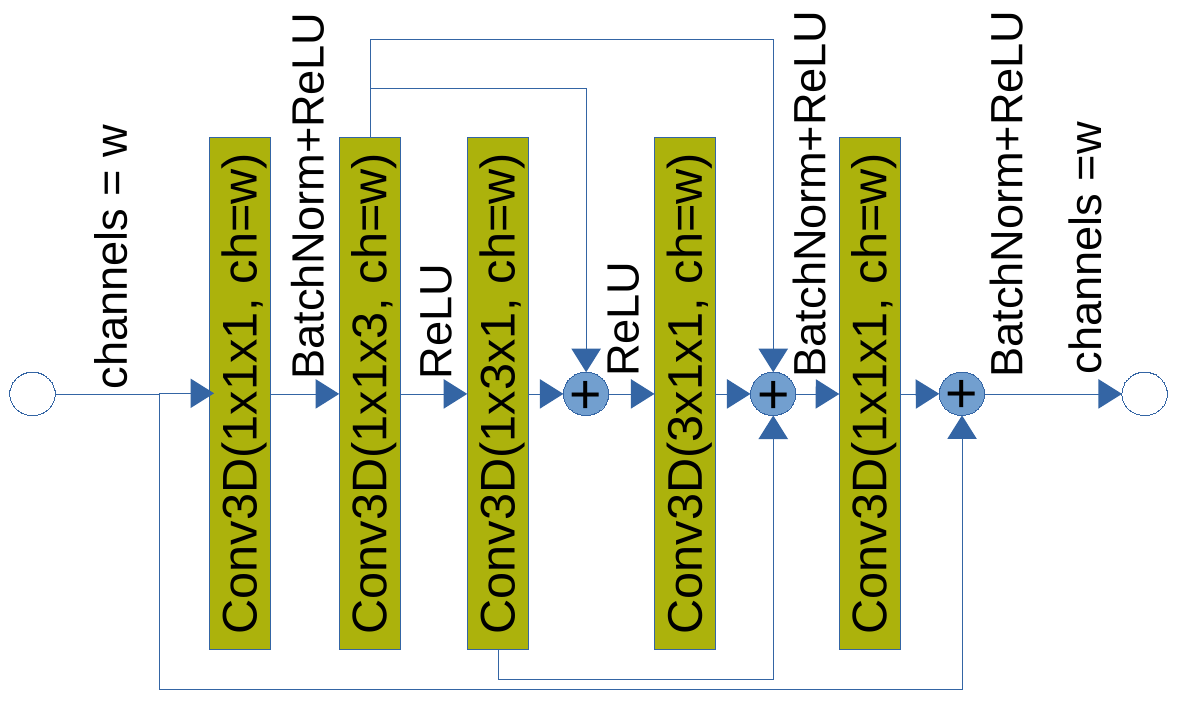}
\caption{\textbf{Proposed \emph{BN-DDR} module}. Our arrangement presents good discrimination and regularization properties while keeping memory consumption manageable.}
\label{fig:BN-DDR}
\end{figure}

\textbf{Batch-normalized DDR.} The fundamental building block of our 3D network is a batch normalized version of the Dimensional Decomposition Residual (DDR) \cite{Li_2019_CVPR} block, named \emph{BN-DDR}. The DDR block is a lightweight alternative to the ResNet block \cite{ResNet} when facing the vanish gradient problem in deep neural networks. Our preliminary experiments showed that adding batch normalization layers to the DDR block produces better results. However, adding a batch normalization layer after each convolutional layer of the block as in \cite{attention_2020}, consumes too much memory, making the network difficult to train with larger mini-batches, reducing the overall benefits of the batch normalization, and making the training slower. Our solution eliminates the batch normalization layers between the dimensional decomposition layers resulting in our proposed  \emph{BN-DDR} block as presented in Figure \ref{fig:BN-DDR}. Due to its reduced memory footprint, we keep the same number of channels of the outer layers in the inner layers, 
while keeping the mini-batch with 4 scenes. For example, previous DDR-based solutions like DDR-Net \cite{Li_2019_CVPR} and AMFNet \cite{attention_2020} use a batch size of 2 and 1, respectively. SketchAware \cite{Sketch} uses a batch size of 4, as well, however it requires 2 11GB GPUs for training.

\textbf{Fusing structure and semantics.} We use an early/late fusion strategy to fuse the detailed structure information from the F-TSDF encoded high-resolution volume to the semantic information from the surface prior volume, as presented in Figure \ref{fig:arch}. The two initial branches are used to match the resolution and number of channels of both inputs. Then, both signals are early fused in an encoder-decoder 3D CNN with a skip connection in the mid-resolution stage, inspired by the U-Net architecture \cite{unet_2015}. This fusion helps to preserve the details of the higher resolution level. The outer skip connections provide additional structural and semantic guidance, and the output branch performs the final late fusion and classification into the desired number of classes.

\textbf{Data balancing and loss function.}
There are two main sources of data unbalancing in the 3D SSC problem: the first is the unbalance between occluded empty and occupied voxels; the second is the unbalance between the several classes. In this work, we face both unbalances in our loss functions. For the first one, we follow \cite{edgnet} and, for each training mini-batch, we randomly sample the occluded voxels to balance occupied and empty voxels, while ignoring empty voxels in the visible space. For the class unbalance, we use a class-weighted cross-entropy loss function, where the weight $w_c = 2$ for less frequent classes like TVs, objects, table and chair and $w_c = 1$ for all other classes.
Being $V$ the set of voxels of the mini-batch ($\textrm{lmb}$) selected for evaluation, $v \in V$,
$n$ the number of classes, $y_{v,c}$ a binary indicator if the voxel $v$ belongs to class $c$ and $P_{v,c}$ the predicted probability of the
voxel $v$ related to class $c$, the loss $L$ is given by equation \ref{eq:2}.
\begin{equation}
L= \frac{1}{|V|}\sum_v \left( - \frac{\sum_{c=1}^{n} [ w_c \cdot y_{v,c} \cdot \log(P_{v,c})]    }{\sum_{c=1}^{n} w_c} \right)
\label{eq:2}
\end{equation}

\subsection{Data Augmentation for SSC}
\begin{figure}[hbt]
\centering  

\begin{subfigure}{0.15\textwidth}
\includegraphics[width=\linewidth]{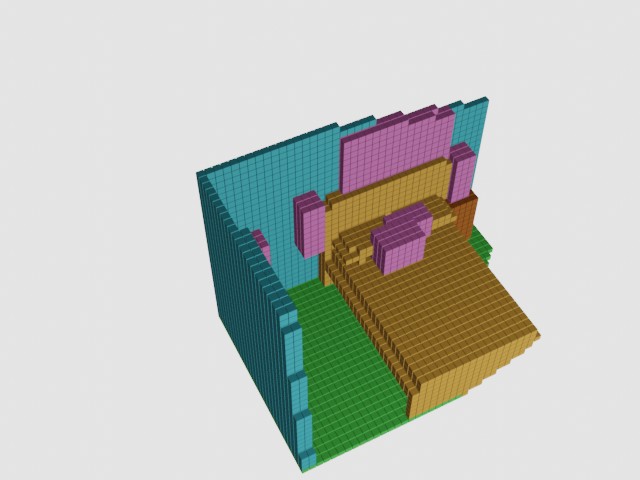}
\vspace*{-6mm}
\caption{no transf.}
\end{subfigure}
\begin{subfigure}{0.15\textwidth}
\includegraphics[trim=0 0 0 0 , clip, width=\linewidth]{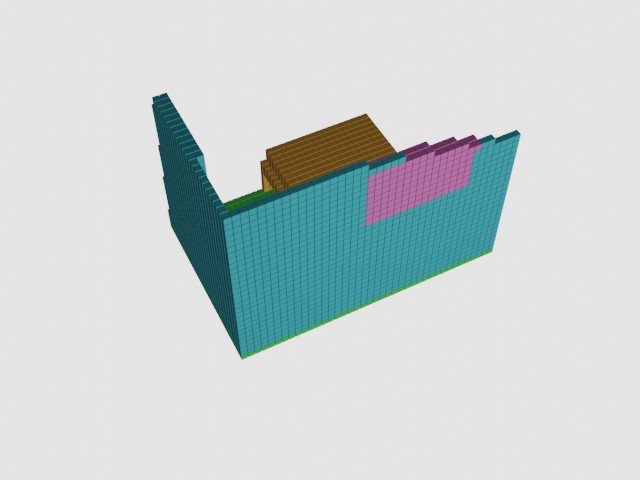}
\vspace*{-6mm}
\caption{t1}
\end{subfigure}
\begin{subfigure}{0.15\textwidth}
\includegraphics[trim=0 0 0 0 , clip, width=\linewidth]{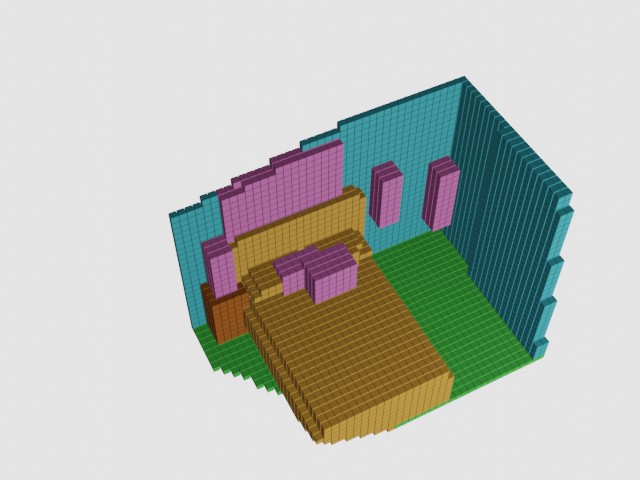}
\vspace*{-6mm}
\caption{t2}
\end{subfigure}
\begin{subfigure}{0.15\textwidth}
\includegraphics[trim=0 0 0 0 , clip, width=\linewidth]{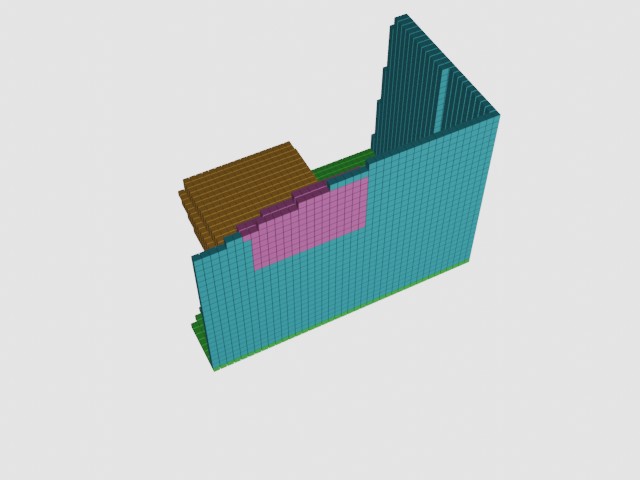}
\vspace*{-6mm}
\caption{t1 and t2}
\end{subfigure}
\begin{subfigure}{0.15\textwidth}
\includegraphics[trim=0 0 0 0 , clip, width=\linewidth]{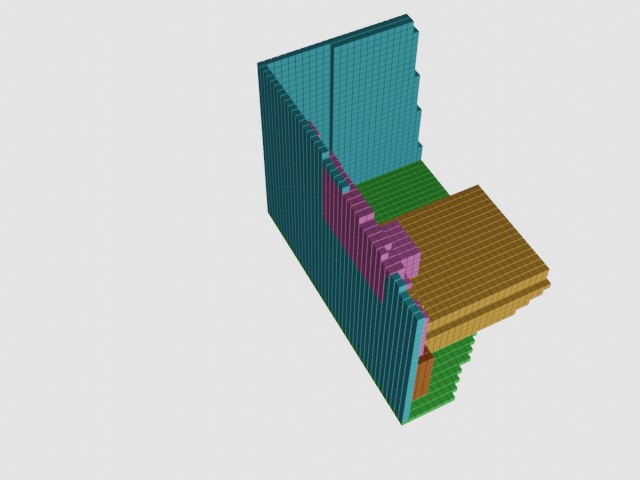}
\vspace*{-6mm}
\caption{t3}
\end{subfigure}
\begin{subfigure}{0.15\textwidth}
\includegraphics[width=\linewidth]{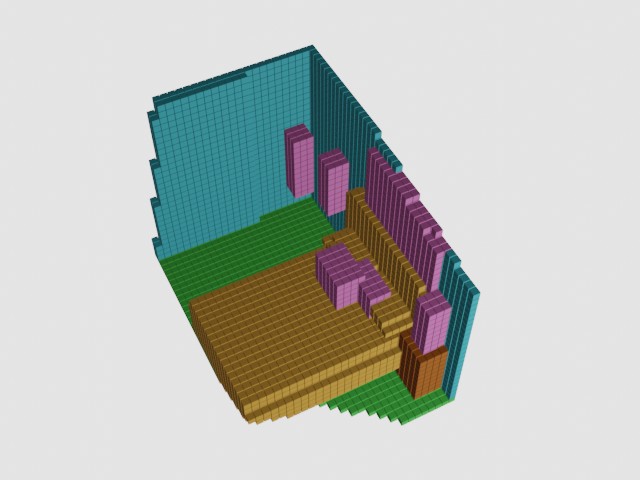}
\vspace*{-6mm}
\caption{t1 and t3}
\end{subfigure}
\begin{subfigure}{0.15\textwidth}
\includegraphics[trim=0 0 0 0 , clip, width=\linewidth]{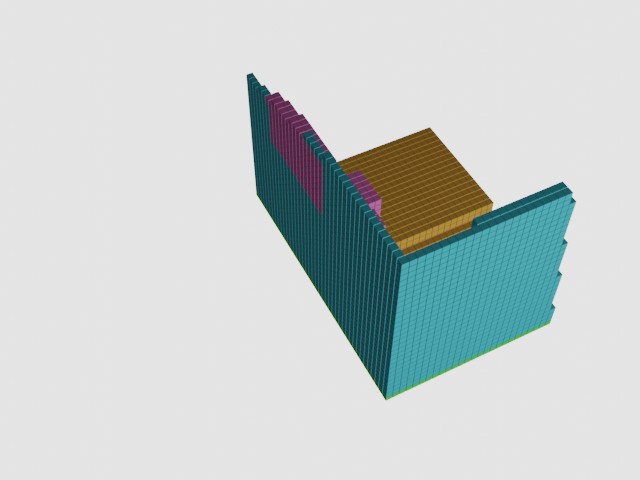}
\vspace*{-6mm}
\caption{t2 and t3}
\end{subfigure}
\begin{subfigure}{0.15\textwidth}
\includegraphics[trim=0 0 0 0 , clip, width=\linewidth]{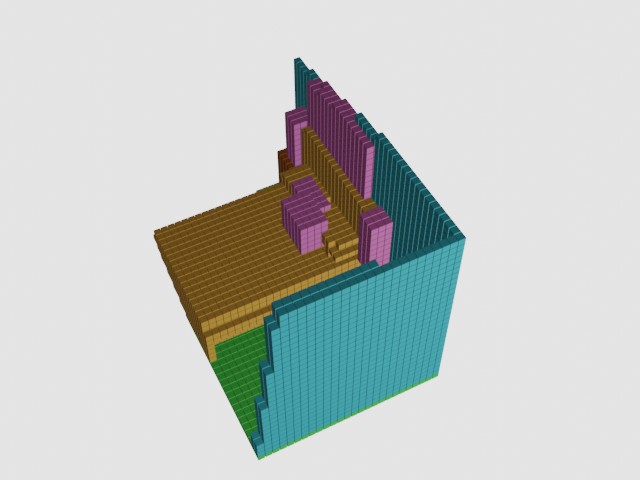}
\vspace*{-6mm}
\caption{t1, t2 and t3}
\end{subfigure}
\caption{\textbf{All augmented volumes generated from a single scene}. Each image caption indicates which transformations were applied. The transformations are: t1 ($\mathrm{X}$-axis flipping), t2 ($\mathrm{Z}$-axis flipping) and t3 ($\mathrm{X}\leftrightarrow\mathrm{Z}$ axes swapping). } 
\label{fig:da}
\end{figure}

Data augmentation techniques have proven to be very effective in 2D segmentation problems, however, there are obstacles that make it difficult to apply them for the problem of  semantic scene completion, where the goal is to map from RGB-D images to 3D labeled voxels. 
To overcome the lack of pixel-to-voxel correspondence in those applications, 
we introduce the use of data augmentation applied directly to the projected 3D volumes of the SSC solutions. We randomly rotate the scene in 45-degree steps and randomly flip along the horizontal axes ($\mathrm{X}$ and $\mathrm{Z}$), to avoid generating upside-down scenes. This precaution is usually taken in 2D domains. For instance, vertical flipping and mirroring are usually avoided in 2D.  We achieve this augmentation with 3 simple and fast 3D transformations. The first two transformations are random $\mathrm{X}$-axis flipping and random $\mathrm{Z}$-axis flipping. The third transformation is random $\mathrm{X}\leftrightarrow\mathrm{Z}$ axes swapping.
The use of this strategy during training is equivalent to augment eight times the size of the dataset. Figure \ref{fig:da} illustrates all possible augmented volumes from a single scene.  All operations are done in the GPU tensors and can be easily executed in parallel, with almost no impact on training time. The proposed data augmentation strategy is mode and resolution agnostic, thus, it may be readily applied to multi-modal or to multi-resolution SSC setups. For instance, SPAwN 3D mini-batches contain the F-TSDF encoded high resolution surface volume and the 12 channels of low resolution priors.  


\textbf{Training-time data augmentation}. During training, for each training mini-batch, we randomly choose to apply or not each one of the 3 transformations. The chosen transformations are then applied to the whole 3D input mini-batch at once and also to the ground truth. 

\textbf{Test-time data augmentation}. We also evaluate the use of the transformations in test-time. In this case, for each test mini-batch, we apply all the eight possible combinations of transformation to each input volumes, generate the predicted volumes and, unlike in training-time, we apply the inverse transformation to the output volumes instead of the same transformation. In this way, the eight output volumes share the same orientation as the original mini-batch. The aligned predictions are then ensembled. For that, we apply the ``sum rule'' \cite{Kittler_1998}, generating a single and more accurate output.

\section{Experiments}

\subsection{Datasets}

We executed our experiments on the three most important SSC benchmark datasets: NYUDv2 \cite{silberman_2012}, NYUCAD \cite{firman_2016} and SUNCG \cite{song_semantic_2017}.

\textbf{NYUDv2} includes depth and RGB images captured by the Kinect depth sensor gathered from commercial and residential buildings, comprising 464 different indoor scenes, divided into 795 samples for training and 654 for testing. We generated ground truth by voxelizing the 3D mesh annotations from \cite{Guo2015} and mapped object categories based on \cite{Handa2015} to label occupied voxels with semantic classes. NYUDv2 is a challenging dataset due to the small number of images for training and the misalignment between ground truth generated from 3D objects meshes and the depth maps.

\textbf{NYUCAD} is a dataset generated from NYUD-v2 where the depth maps are synthesized from the 3D objects meshes, eliminating the misalignment between the depth map and the 3D ground truth. The RGB images and 2D segmentation ground truth are the original ones from NYUDv2. Hence, they present misalignment when compared to the generated depth map and the 3D ground truth. To reduce the effect of this remaining misalignment on 2D modes, when training our 2D segmentation network, we use surface normals extracted from the synthetic depth maps, and the 2D ground truth is re-generated back-projecting the 3D ground truth to 2D. 

\textbf{SUNCG} dataset consists of about 45K completely synthetic house models from which were extracted more than 130K 3D snapshots with corresponding depth maps and ground truths divided into train and test datasets. As SUNCG neither includes the RGB images nor the surface normals, we render the RGB images using the provided camera positions for each snapshot as specified in~\cite{edgnet} and extract the surface normals from the depth map. 

\subsection{Training Details}
All models of the experiments conducted in this paper were optimized with SGD using the \emph{one cycle learning rate policy} \cite{Smith_2018} with the maximum and minimum learning rates (LR) multipliers set to 25 and $1 \times 10^{-4}$, respectively, cosine annealing, $1 \times 10^{-5}$ weight decay, and mini-batches with four scenes. 

\textbf{2D network training.} 2D models were trained in two data augmented stages. In the first stage, the ImageNet pretrained ResNet-101 backbones weights were frozen, and the base LR was set to $2 \times 10^{-4}$. In the second stage, the base LR was set to $8 \times 10^{-5}$, and all weights were unfrozen, but the ResNet backbones LR was set to 1/10 of the running LR. For NYUDv2 and NYUCAD, each stage comprises 150 epochs, and for SUNCG, 10 epochs. The data augmentation transforms were the conventional 2D random resize, random crop, and random horizontal flip.

\textbf{3D network training.} 3D models were trained in a single stage, with base LR set to $4 \times 10^{-4}$. 
Our 3D data augmentation strategy reduces overfitting and thus, allows us to train for more epochs. For NYUDv2 and NYUCAD, the models were trained for 80 epochs when not using data augmentation and for 120 epochs when augmented. As SUNCG is a very large dataset, the benefits of data augmentation is expected to be inexpressive, so we do not apply data augmentation on SUNCG and train for only 10 epochs.

\textbf{Fine-tuning from SUNCG.} When fine-tuning from SUNCG, both 2D and 3D models were first trained on SUNCG then fine-tuned to the desired target dataset. The same protocols described previously were applied.

\textbf{Metrics.}
We follow the previous works and report scores for the completion task and the semantic scene completion task. For completion we report precision, recall and Intersection over Union (IoU) considering the prediction of occupancy for each occluded voxel. For the semantic scene completion task we report the IoU of each object class on both the observed and occluded voxels and the averaged result over all classes except the void class (mIoU). 
Voxels outside the view and visible empty voxels are ignored.
\subsection{Ablation Study}

We evaluate the importance of each aspect of the proposed solution with a comprehensive set of experiments using only real images from NYUDv2 without pretraining on synthetic images.
Table \ref{tab:ablation} presents the progressive contribution of the proposed \emph{BN-DDR} module, the use of the proposed class balancing strategy, and the data augmentation training and test approaches, considering one, two, and three input modes. All models were trained and evaluated under the same protocols except for the number of epochs when using data augmentation, as explained in previous section.


We observed positive contributions for each of the evaluated aspects. Using RGB and surface normals as inputs consistently produced positive impacts. The model itself surpassed state-of-the-art-results with regular not augmented training. Moreover, combining the proposed network with the data augmentation approaches enhances results to unprecedented levels.

\begingroup
\setlength{\tabcolsep}{2pt} 
\renewcommand{\arraystretch}{1} 

\begin{table}
  \centering
  \begin{tabular}
      { c| c | c | c | c |c | c }
  \hline
     \multirow{2}{*}{\shortstack{input\\ modes}} & DDR & {class}  & \multirow{2}{*}{DA} & \multirow{2}{*}{TTDA} & comp. & {SSC}\\
         & type   & bal.  & &  & IoU & mIoU  \\
  \midrule\midrule
        \multirow{3}{*}{depth} & Regular & no & no & no & 55.5 & 24.5\\
         & \emph{BN-DDR} & no  &  no & no & 60.8 & 31.8\\\
         & \emph{BN-DDR} & yes  & no & no &  60.8 & 32.2\\
 \hline
         \multirow{3}{*}{\shortstack{depth\\ rgb}} & Regular & no  &  no & no & 60.9 & 38.6\\
         & \emph{BN-DDR} & no   & no & no & 63.0 & 41.0\\
         & \emph{BN-DDR} & yes   & no & no & 64.4& 42.2\\
 \hline
        \multirow{5}{*}{\shortstack{depth\\ rgb\\ sn}} & Regular & no  & no & no & 61.3 & 39.2\\
         & \emph{BN-DDR} & no   & no & no & 63.4 & 41.4\\
         & \emph{BN-DDR} & yes   & no & no & 63.8 & 43.4\\
         & \emph{BN-DDR} & yes   & yes & no & 65.7 & 47.7\\
         & \emph{BN-DDR} & yes   & yes & yes & 66.2 & 48.0\\

 \midrule\midrule
        {oracle test} & \emph{BN-DDR} & yes &  no & no & 76.7 & 67.9\\
 \hline
   
  \end{tabular}
  \caption{\textbf{Progressive impact of SPAwN components on NYUDv2.} No pretraining was performed. ``sn'' means surface normals, DA means data augmentation, and TTDA means test-time data augmentation.}
  \label{tab:ablation}
\end{table}
\endgroup

\begingroup
\setlength{\tabcolsep}{2pt} 
\renewcommand{\arraystretch}{1} 
\begin{table*}[ht!]
  \centering
  \begin{tabular}
            {  c | c | c c c | c c c c c c c c c c c c}
      \hline
     \multirow{ 2}{*}{model} & \multirow{ 2}{*}{\shortstack{pipeline\\type}} & \multicolumn{3}{c|}{scene completion} & \multicolumn{12}{c}{semantic scene completion (IoU, in percentages)} \\
    & & prec. & rec. & IoU & ceil. & floor & wall & win. & chair & bed & sofa & table & tvs & furn. & objs. & avg. \\
    
    \hline

    SISNet-BiSeNet \cite{Cai_2021_CVPR} & \multirow{2}{*}{iterative}
    & 93.3 & 96.1 & 89.9 & 85.2 & 90.0 & 83.7 & 80.8 & 60.0 & 83.5 & 80.8& 68.6 & 77.3 & 86.7 & 70.1 & 78.8  \\

    SISNet-DeepLabv3 \cite{Cai_2021_CVPR}  
    && 92.6 & 96.3 & 89.3 & 85.4 & 90.6 & 82.6 & 80.9 & 62.9 & 84.5 &  82.6 & 71.6 & 72.6 & 85.6 & 69.7 & 79.0   \\

    \midrule\midrule

EdgeNet\cite{edgnet} & \multirow{ 4}{*}{\shortstack{straight-\\forward}} & \underline{93.3} & 90.6 & \underline{85.1}  & 97.2 & \underline{95.3} & \underline{78.2} & 57.5,
    & 51.4 & 80.7 & 74.1 & 54.5 & 52.6 & 70.3 &  60.1 & 70.2\\

    ESSC\cite{Zhang_2018_ECCV} && 92.6 & 90.4 & 84.5& 96.6 & 83.7 & 74.9 & 59.0 & 55.1 & \underline{83.3} & 78.0 & 61.5 & 47.4 & 73.5 & 62.9 & 70.5\\
    
    CCPNet\cite{CCPNet} &&\textbf{98.2}&\textbf{96.8}&\textbf{91.4}&\underline{99.2}&89.3&76.2&\underline{63.3}&\underline{58.2}&\textbf{86.1}&\textbf{82.6}&\underline{65.6}&\underline{53.2}&\textbf{76.8}&\underline{65.2}&\underline{74.2}\\
    
       \textbf{SPAwN} (ours) & 
       & 91.9 & 88.7 & 82.3 & \textbf{99.3} & \textbf{96.1} & \textbf{84.4} & \textbf{75.1} & \textbf{59.2} & 81.5 & \underline{78.1} & \textbf{67.3} & \textbf{80.1}  & \underline{76.3} & \textbf{70.4} & \textbf{78.9} \\ 
\hline
  \end{tabular}
  \caption{\textbf{Results on SUNCG test set}. 
  Our SPAwN semantic scene completion overall results surpass by far all known previous straight-forward solutions on SUNCG synthetic images, and are comparable to both SISNet models, even though they have a much higher parameter count and operate with a complext iterative pipeline for both training and inference. 
  }
  \label{tab:res_suncg}
\end{table*}
\endgroup


\begingroup
\setlength{\tabcolsep}{2pt} 
\renewcommand{\arraystretch}{1} 

\begin{table*}
  \centering
  \begin{tabular}[h!]
      { c | c | c | c c c | c c c c c c c c c c c c}
  \hline
      \multirow{ 2}{*}{model} &
      \multirow{ 2}{*}{\shortstack{pipeline\\type}} &
      \multirow{ 2}{*}{train}& 
      \multicolumn{3}{c|}{scene compl.}& \multicolumn{12}{c}{semantic scene completion (IoU, in percentages)} \\
        & & & prec. & rec. & IoU & ceil. & floor & wall & win. & chair & bed & sofa & table & tvs & furn. & objs. & avg. \\
    \midrule\midrule

   SISNet-BiSeNet\cite{Cai_2021_CVPR}  & 
   \multirow{ 2}{*}{iterative}& \multirow{ 2}{*}{NYU} &
   90.7 & 84.6 & 77.8 & 53.9 & 93.2 & 51.3 & 38.0 & 38.7 & 65.0 & 56.3 & 37.8 & 25.9 & 51.3 & 36.0 & 49.8  \\

   SISNet-DLabv3\cite{Cai_2021_CVPR}  & 
   &  &
   92.1 & 83.8 & 78.2 & 54.7 & 93.8 & 53.2 & 41.9 & 43.6 & 66.2 & 61.4 & 38.1 & 29.8 & 53.9 & 40.3 & 52.4  \\

\midrule\midrule

    TS3D\cite{Garbade_2019_CVPR_Workshops}&
    \multirow{ 3}{*}{\shortstack{straight-\\forward}}&\multirow{ 3}{*}{NYU} &-&-&
    60.0 & 9.7& 93.4& 25.5&21.0& 17.4& 55.9& 49.2& 17.0& \underline{27.5}& 39.4 &19.3 &34.1\\

    SketchAware\cite{Sketch} & 
    & {NYU} &
    \textbf{85.0} & \textbf{81.6} & \textbf{71.3} &\textbf{43.1} &\underline{93.6} &\textbf{40.5} &\underline{24.3} &\underline{30.0} &\underline{57.1} &\underline{49.3} &\underline{29.2} &14.3 &\underline{42.5} &\underline{28.6} &\underline{41.1}\\

    {\textbf{SPAwN} (ours)} & 
    &  &
    \underline{82.3} & \underline{77.2} & \underline{66.2} & \underline{41.5} & \textbf{94.3} & \underline{38.2} & \textbf{30.3} & \textbf{41.0} & \textbf{70.6} & \textbf{57.7} & \textbf{29.7} & \textbf{40.9} & \textbf{49.2} & \textbf{34.6} & \textbf{48.0} \\

     \midrule\midrule
    TNetFuse\cite{See_and_think_2018} & \multirow{ 4}{*}{\shortstack{straight-\\forward}}
    &\multirow{ 4}{*}{\shortstack{NYU\\+\\SUNCG}} &
    67.3 & \underline{85.8} & 60.6& 17.3 & 92.1 & 28.0 & 16.6 & 19.3 & 57.5 & 53.8 & 17.7 & 18.5 & 38.4 & 18.9 & 34.4\\

    ForkNet\cite{ForkNet}&
    &&
    -&-&63.4& \underline{36.2}& 93.8 &29.2 &18.9& 17.7& 61.6& 52.9 &\underline{23.3} &19.5& \underline{45.4} &20.0 &37.1\\

    CCPNet\cite{CCPNet}& 
    & &
    \textbf{91.3} & \textbf{92.6} & \textbf{82.4}& 25.5 &\textbf{98.5}& \underline{38.8}& \underline{27.1} &\underline{27.3} &\underline{64.8} &\textbf{58.4} &21.5 &\underline{30.1} &38.4 &\underline{23.8} &\underline{41.3}\\

       {\textbf{SPAwN} (ours)} & &  &      
      \underline{81.2} & 80.4 & \underline{67.8} & \textbf{44.2} & \underline{94.2} & \textbf{40.9} & \textbf{33.5} & \textbf{42.5} & \textbf{69.3} & \textbf{58.4} & \textbf{32.4} & \textbf{44.3} & \textbf{53.4} & \textbf{36.3} & \textbf{49.9}  \\
      
       \hline
  \end{tabular}
  \caption{\textbf{Results on NYUDv2 test set}. SUNCG + NYU means trained on SUNCG and fine-tuned on NYUDv2.
  Our SPAwN  models hold the best and second-best overall semantic scene completion results for real-world images, on both training scenarios, when compared to previous straight-forward solutions.}
  \label{tab:res_nyu}
\end{table*}
\endgroup


\begingroup
\setlength{\tabcolsep}{2pt} 
\renewcommand{\arraystretch}{1} 

\begin{table*}
  \centering
  
    \begin{tabular}[h!]
      { c | c | c | c c c | c c c c c c c c c c c c}
  \hline
      \multirow{ 2}{*}{model} &
      \multirow{ 2}{*}{\shortstack{pipeline\\type}} &
      \multirow{ 2}{*}{train}& 
      \multicolumn{3}{c|}{scene compl.}& \multicolumn{12}{c}{semantic scene completion (IoU, in percentages)} \\
        & & & prec. & rec. & IoU & ceil. & floor & wall & win. & chair & bed & sofa & table & tvs & furn. & objs. & avg.  \\
    \midrule\midrule

   SISNet-BiSeNet\cite{Cai_2021_CVPR}  & 
   \multirow{ 2}{*}{iterative}& \multirow{ 2}{*}{NYUCAD} &
   94.2 & 91.3 & 86.5 & 65.6 & 94.4 & 67.1 & 45.2 & 57.2 & 75.5 & 66.4 & 50.9 & 31.1 & 62.5 & 42.9 & 59.9  \\

   SISNet-DLabv3\cite{Cai_2021_CVPR}  & 
   & & 
   94.1 & 91.2 & 86.3 & 63.4 & 94.4 & 67.2 & 52.4 & 59.2 & 77.9 & 71.1 & 58.1 & 46.2 & 65.8 & 48.8 & 63.5  \\

\midrule\midrule

    CCPNet\cite{CCPNet}& \multirow{ 3}{*}{\shortstack{straight-\\forward}}
    &\multirow{ 3}{*}{\shortstack{NYUCAD}} &
    \textbf{91.3} & \textbf{92.6} & \underline{82.4} & 56.2& \textbf{96.6}& 58.7& \underline{35.1}& 44.8& 68.6& 65.3& 37.6 &\underline{35.5}& 53.1& 35.2& 53.2\\

    SketchAware\cite{Sketch} & 
    & &
    \underline{90.6} & \underline{92.2} & \textbf{84.2} &\underline{59.7} &94.3 &\textbf{64.3} &32.6&\underline{51.7} &\underline{72.0} &\underline{68.7} &\underline{45.9} &19.0 &\underline{60.5} &\underline{38.5} &\underline{55.2}\\

    {\textbf{SPAwN (ours)} } & & & 
    {84.5} & 87.8 & {75.6} & \textbf{65.3} & \underline{94.7} & \underline{61.9} & \textbf{36.9} & \textbf{69.6} &\textbf{82.2} & \textbf{72.8} & \textbf{49.1} & \textbf{43.6} & \textbf{63.4} & \textbf{44.4} & \textbf{62.2} \\

     \midrule\midrule
    SSCNet\cite{song_semantic_2017} & \multirow{ 3}{*}{\shortstack{straight-\\forward}}
    &\multirow{ 3}{*}{\shortstack{NYUCAD\\+\\SUNCG}} &
    75.4 & \textbf{96.3} & 73.2 & 32.5 & 92.6 & 40.2 & 8.9 & 40.0 & 60.0 & 62.5 & 34.0 & 9.4 & 49.2 & 26.5 & 40.0\\

    CCPNet\cite{CCPNet}& 
    & &
    \textbf{93.4} & \underline{91.2} & \textbf{85.1} & \underline{58.1} &\textbf{95.1}& \underline{60.5}& \underline{36.8} &\underline{47.2} &\underline{69.3} &\underline{67.7} &\underline{39.8} &
      \underline{37.6} &\underline{55.4} &\underline{37.6} &\underline{55.0}\\

       {\textbf{SPAwN} (ours)} & &  &      
         86.3 & 90.1 & \underline{78.9} & \textbf{77.6} & \underline{95.0} & \textbf{68.0} & \textbf{38.1} & \textbf{67.9} & \textbf{82.2} & \textbf{77.1} & \textbf{56.8} & \textbf{50.0} & \textbf{65.7} & \textbf{46.5} & \textbf{65.9} \\
      
       \hline
  \end{tabular}
  
    \caption{\textbf{Results on NYUDCAD}. 
    Our SPAwN  models hold the best and second-best overall results on both training scenarios, when compared to previous straight-forward solutions. When fine-tuned from SUNCG, SPAwN surpasses both SISNet models, which are much more complex than ours. }
  \label{tab:res_nyucad}
\end{table*}
\endgroup

Table \ref{tab:ablation} also evaluates the model's theoretical upper bound limit in an Oracle Test, supposing we have predicted perfect semantic 2D priors. To this matter, we replace the output of the 2D network with the 2D ground truth. The Oracle experiment shows there is still room for improvements by enhancing 2D predictions. Future works can exploit this.

\begin{figure*}[ht!]
\centering  

\begin{subfigure}{\textwidth}
  \includegraphics[trim=10 240 5 250, clip, width=\linewidth]{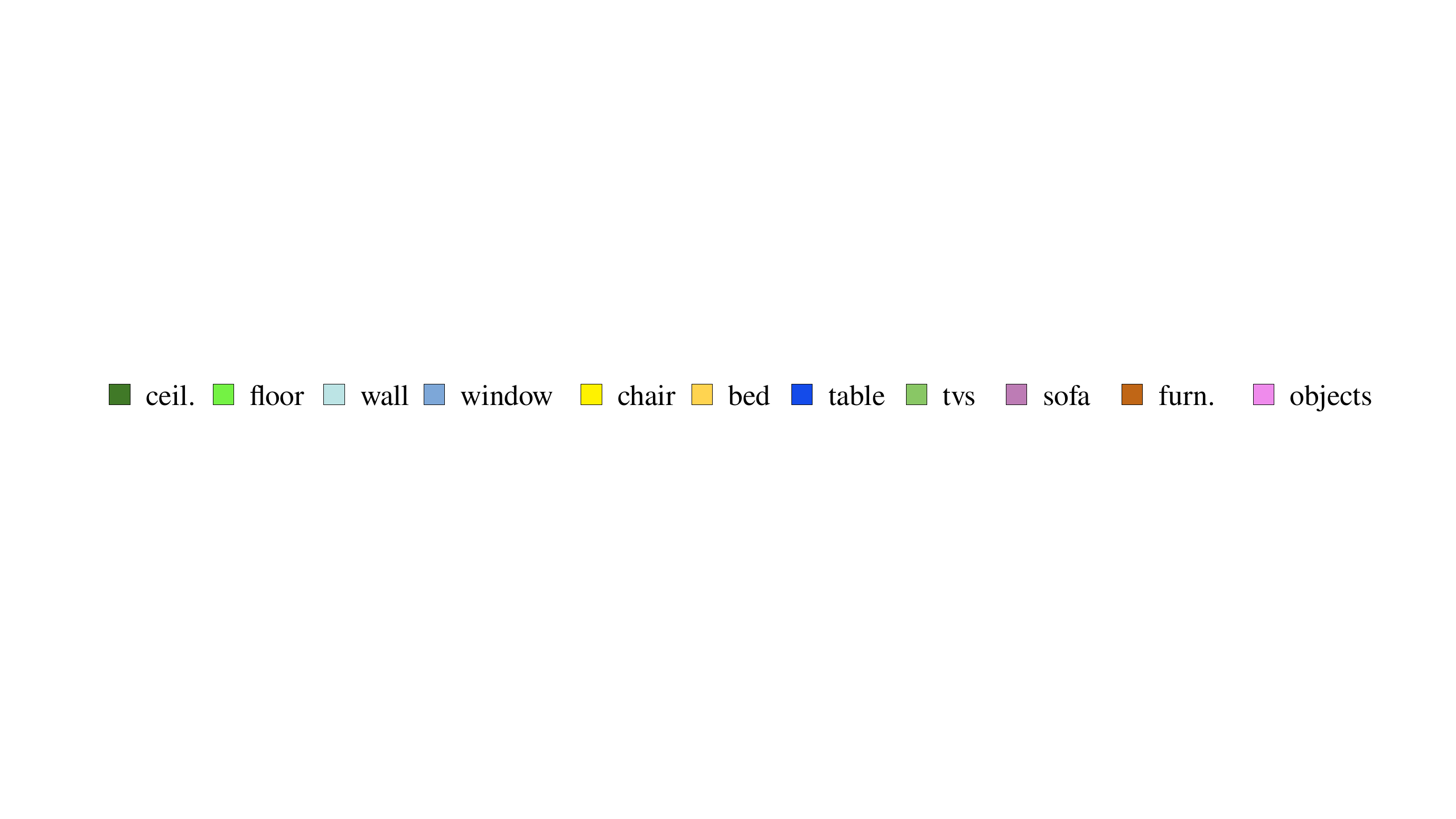}
  
  \end{subfigure}

\bigskip

\begin{subfigure}{0.160\textwidth}
\includegraphics[width=\linewidth]{figures/NYU0190_0000_color.jpg}

\end{subfigure}
\begin{subfigure}{0.160\textwidth}
\includegraphics[trim=0 0 0 0, clip, width=\linewidth]{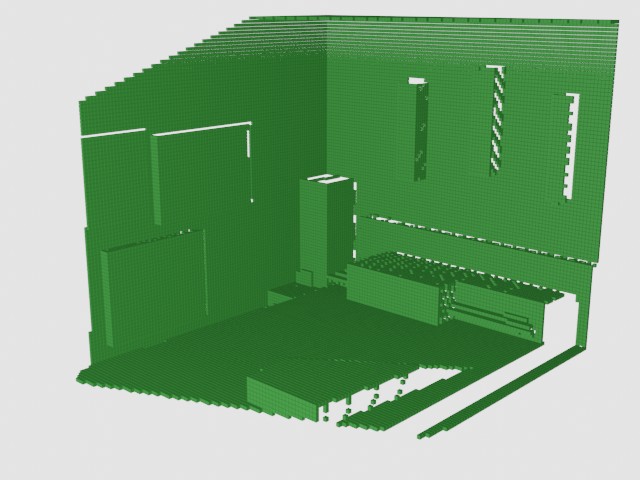}
\end{subfigure}
\begin{subfigure}{0.160\textwidth}
\includegraphics[trim=0 0 0 0, clip, width=\linewidth]{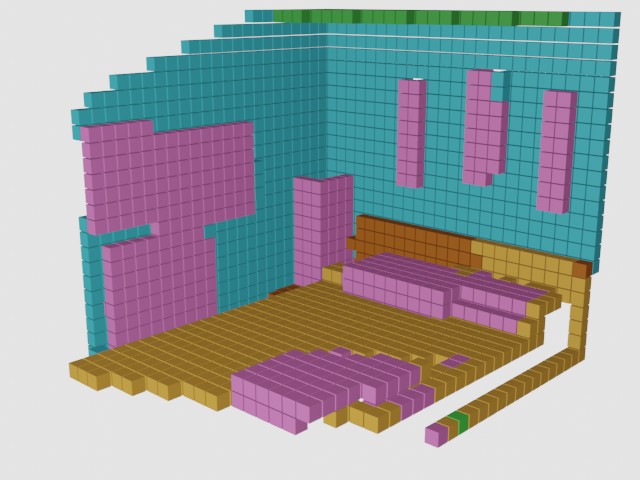}

\end{subfigure}
\begin{subfigure}{0.160\textwidth}
\includegraphics[trim=0 0 0 0, clip, width=\linewidth]{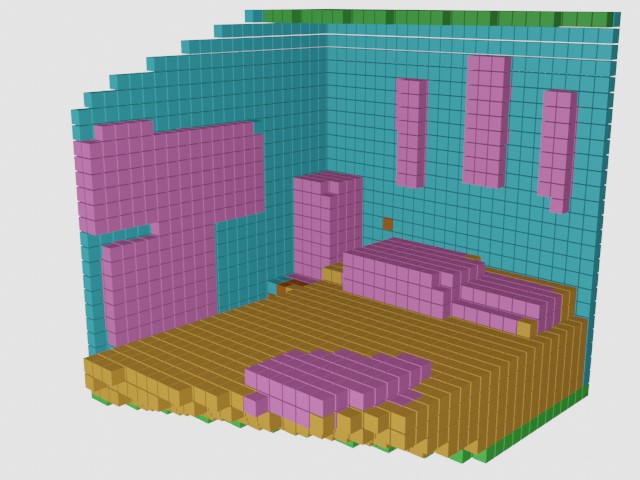}

\end{subfigure}
\begin{subfigure}{0.160\textwidth}
\includegraphics[trim=0 0 0 0, clip, width=\linewidth]{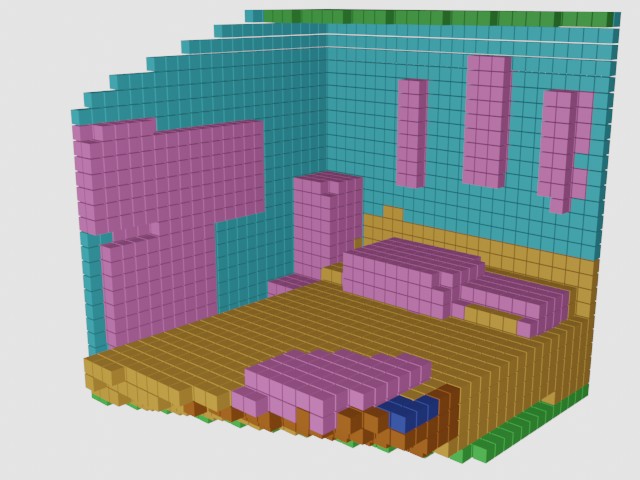}

\end{subfigure}
\begin{subfigure}{0.160\textwidth}
\includegraphics[trim=0 0 0 0, clip, width=\linewidth]{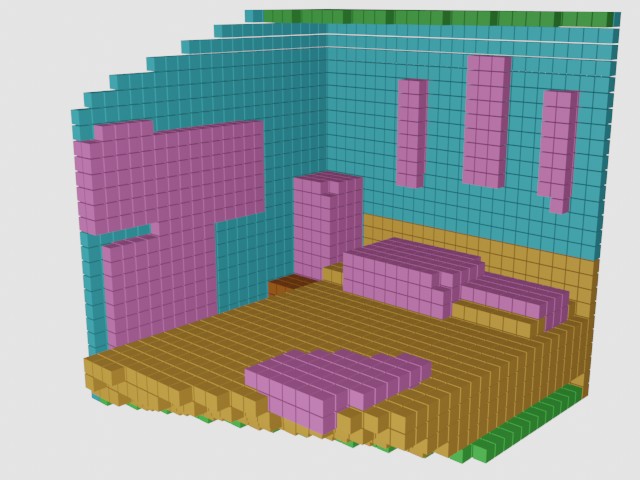}

\end{subfigure}


\begin{subfigure}{0.160\textwidth}
\includegraphics[width=\linewidth]{figures/NYU0310_0000_color.jpg}

\end{subfigure}
\begin{subfigure}{0.160\textwidth}
\includegraphics[trim=0 0 0 0, clip, width=\linewidth]{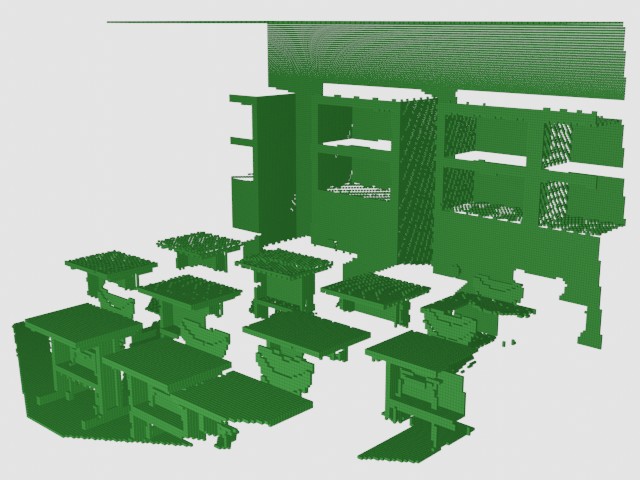}

\end{subfigure}
\begin{subfigure}{0.160\textwidth}
\includegraphics[trim=0 0 0 0, clip, width=\linewidth]{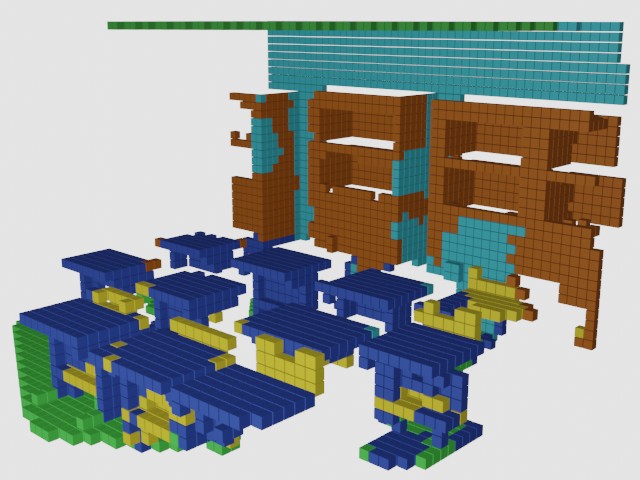}

\end{subfigure}
\begin{subfigure}{0.160\textwidth}
\includegraphics[trim=0 0 0 0, clip, width=\linewidth]{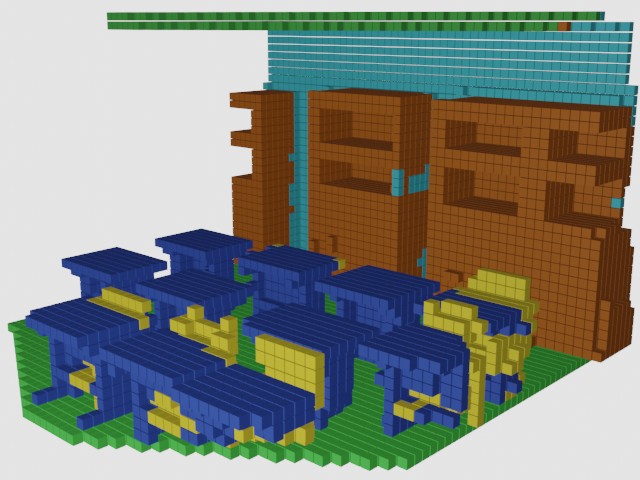}

\end{subfigure}
\begin{subfigure}{0.160\textwidth}
\includegraphics[trim=0 0 0 0, clip, width=\linewidth]{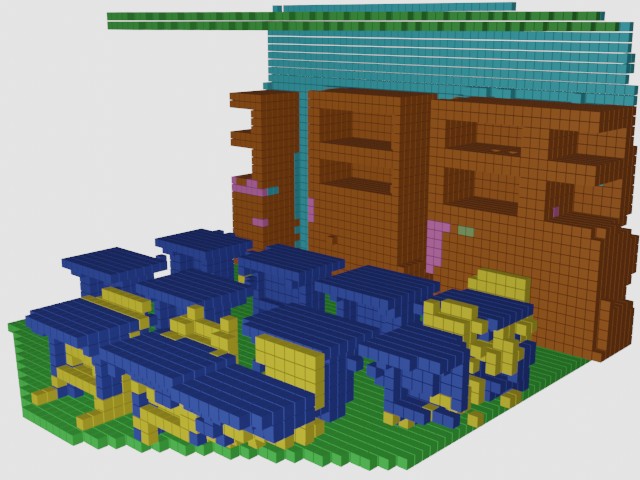}

\end{subfigure}
\begin{subfigure}{0.160\textwidth}
\includegraphics[trim=0 0 0 0, clip, width=\linewidth]{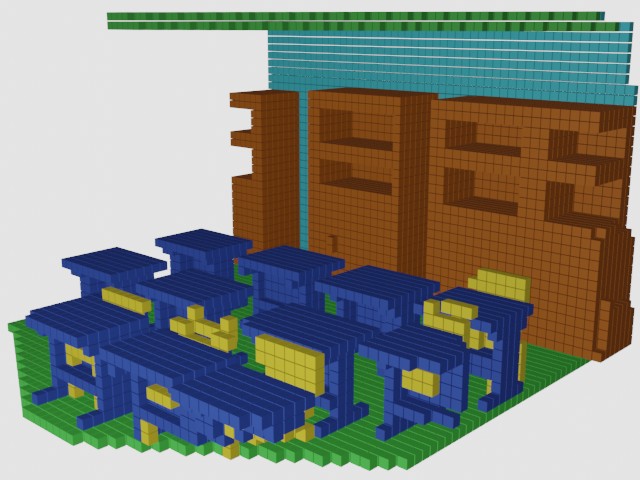}

\end{subfigure}
\begin{subfigure}{0.160\textwidth}
\includegraphics[width=\linewidth]{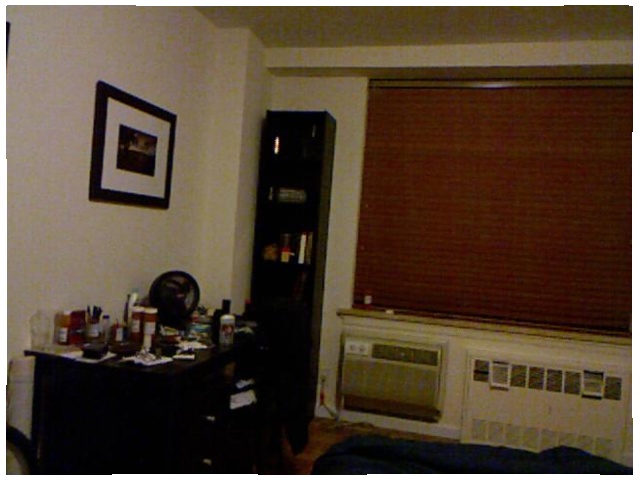}
\caption{RGB}
\label{fig:rgb}

\end{subfigure}
\begin{subfigure}{0.160\textwidth}
\includegraphics[trim=0 0 0 0, clip, width=\linewidth]{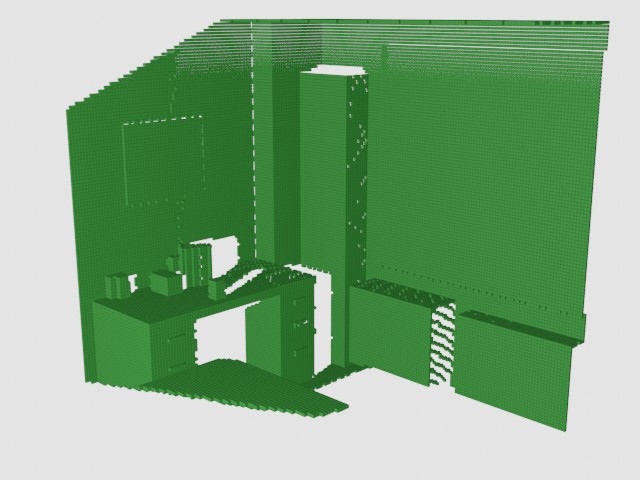}
\caption{Visible surface}
\label{fig:surf}

\end{subfigure}
\begin{subfigure}{0.160\textwidth}
\includegraphics[trim=0 0 0 0, clip, width=\linewidth]{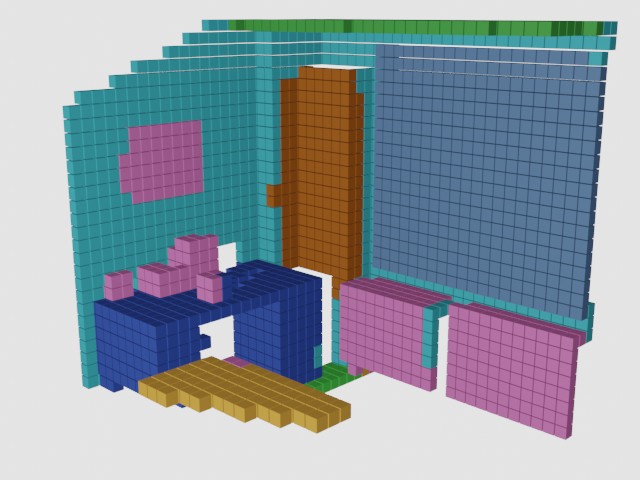}
\caption{Semantic priors}
\label{fig:priors}

\end{subfigure}
\begin{subfigure}{0.160\textwidth}
\includegraphics[trim=0 0 0 0, clip, width=\linewidth]{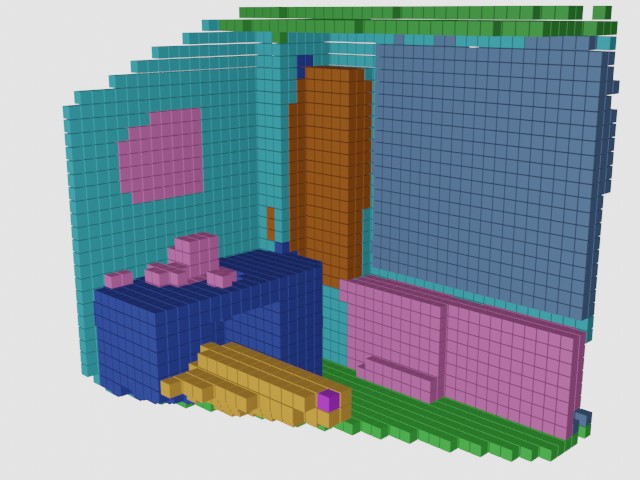}
\caption{SPAwN}
\label{fig:pred}

\end{subfigure}
\begin{subfigure}{0.160\textwidth}
\includegraphics[trim=0 0 0 0, clip, width=\linewidth]{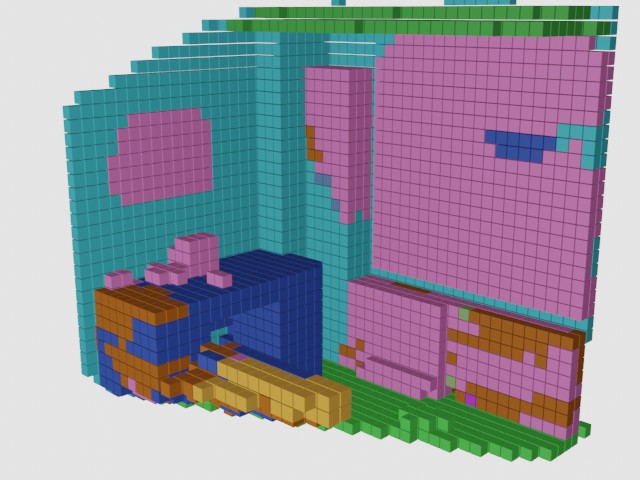}
\caption{SSCNet}
\label{fig:pred2}

\end{subfigure}
\begin{subfigure}{0.160\textwidth}
\includegraphics[trim=0 0 0 0, clip, width=\linewidth]{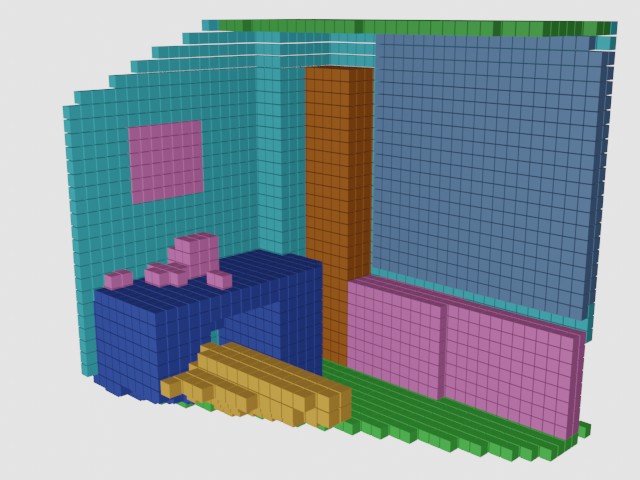}
\caption{GT}
\label{fig:gt}

\end{subfigure}
\caption{\textbf{ SPAwN qualitative results on NYUCAD.} 2D segmentation priors projected to 3D provide good semantic guidance
while SPAwN complete and refine the predictions, achieving results visually close to perfection. Compared to baseline SSCNet \cite{song_semantic_2017}, results are much more accurate. (Best viewed in color).}
\label{fig:qualy}
\end{figure*}

We provide an investigation on the benefits of data augmented  training with real images (NYUDv2) in more detail in supplementary material.

\subsection{Comparison to the State-of-the-Art}
In Tables \ref{tab:res_suncg}, \ref{tab:res_nyu}, and \ref{tab:res_nyucad}, 
for each training scenario, the best scores for straightforward solutions are presented in bold, 
while the second-best scores are underlined. 
We only show the best two or three competing models in each category. Further results are in the supplementary material.

\textbf{SUNCG.} Table \ref{tab:res_suncg} shows the results on SUNCG synthetic dataset. As the SUNCG training set is large, the benefits of data augmentation  are not expected to be significant. Therefore, we only evaluate the standard training  approach. Our proposed 3D CNN surpassed CCPNet, the previous best straightforward solution, by a 6.3\% margin (\SI{4.7}{\pp}) and got similar results to the iterative models.

\textbf{NYUDv2.} Table \ref{tab:res_nyu} presents the results on real images. We evaluated two training scenarios: training from scratch on NYUDv2 and training on SUNCG, then fine-tuning on NYUDv2. Data augmented \emph{SPAwN} presented the best overall results in all scenarios. Without fine-tuning, the boost over SketchAware was  16.7\% (\SI{6.9}{\pp}). With fine-tuning, the bost over CCPNet was 20.8\%  (\SI{8.6}{\pp}) and the result is comparable to the more expensive SISNet models.

\textbf{NYUCAD.} Table \ref{tab:res_nyucad} confirms our expectation of good results due to the better quality of the ground truth related to surface volume and 2D priors. The observed $\textrm{mIoU}$ boost of our model over SketchAware and CCPNet, the best previous solution in each training scenario, is 12.7\% (\SI{7.0}{\pp}) and 19.8\% (\SI{10.9}{\pp}), respectively. Our fine-tuned model even surpassed both expensive SISNet models.

\subsection{Qualitative Analysis}
Figure \ref{fig:qualy} presents a qualitative analysis on NYUCAD due to its better alignment between depth map and ground truth compared to NYUDv2, making it easier to perceive our predictions' high quality visually. We generated predictions with \emph{SPAwN} trained on SUNCG and fine-tuned on NYUCAD, using both training-time and test-time data augmentation. We also qualitatively compare our results to a baseline SSCNet model, pretrained on SUNCG and fine-tuned to NYUCAD. 
This SSCNet model was trained with our one cycle learning rate protocol and achieves better results than the presented in the original paper \cite{song_semantic_2017} (53.3\% vs 40.0\% avg.\ IoU, respectively).
\emph{SPAwN} overall results are perceptible better than SSCNet. In column (c), it is possible to see that our projection and ensemble methods provide first-rate priors to the visible surface, with minimal prediction errors. \emph{SPAwN} fusion strategy can complete the predictions and fix errors from priors, achieving remarkable final results.
Flat objects on flat surfaces are difficult to be detected by depth-only approaches like SSCNet. Notice in the third row of figure \ref{fig:qualy} that SSCNet was unable to identify the window, while \emph{SPAwN} predicitions are almost perfect.
Supplementary material presents qualitative results on the other datasets.

\section{Conclusions}
\vspace{-.1em}
We introduce \emph{SPAwN}, a novel 3D SSC network that explicitly fuses semantic priors with high-resolution structural information from depth maps.
\emph{SPAwN} uses as fundamental building block a
novel lightweight batch normalized DDR module with higher discrimination power than its predecessors. 
We also introduce a the use of 3D data augmentation to the SSC task. The proposed data augmentation strategy is mode and resolution agnostic and may be applied to other SSC solutions. 
An ablation study with a comprehensive set of experiments shows the effectiveness of each one of our contributions. That study also includes an oracle test, which showed that the proposed solution can be further enhanced using better sources of semantic priors.

Data augmented \emph{SPAwN} surpasses by far all previous state-of-the-art solutions with similar complexity in SSC benchmarks, in all training scenarios, achieving a boost of 19.8\% (\SI{10.9}{\pp}) over the best previously reported result on real images. Comparing to the recently introduced and much more expense iterative solution, the improvement is of 3.8\% (\SI{2.4}{\pp}).

\ifwacvfinal

\section*{Acknowledgments}
Dr.\ de Campos acknowledges the support of CNPq fellowship PQ 314154/2018-3 and FAPDF project KnEDLe, convênio 07/2019. He currently works at Vicon Motion Systems, Oxford Metrics Group, UK.
Mr.\ Dourado thanks TCU (Tribunal de Contas da União) for supporting his research.
\else
\fi

{\small
\bibliographystyle{ieee_fullname}
\bibliography{paper}
}

\end{document}